\documentclass[sigconf]{aamas} 

\usepackage{balance} 
\usepackage{algorithm} 
\usepackage{algpseudocode}
\usepackage{xcolor}

\setcopyright{ifaamas}
\acmConference[AAMAS '23]{Proc.\@ of the 22nd International Conference
on Autonomous Agents and Multiagent Systems (AAMAS 2023)}{May 29 -- June 2, 2023}
{London, United Kingdom}{A.~Ricci, W.~Yeoh, N.~Agmon, B.~An (eds.)}
\copyrightyear{2023}
\acmYear{2023}
\acmDOI{}
\acmPrice{}
\acmISBN{}

\acmSubmissionID{1049}

\title[AAMAS-2023 Monopoly]{Methods and Mechanisms for Interactive Novelty Handling in Adversarial Environments }

\author{Tung Thai}
\affiliation{
  \institution{Tufts University}
  \city{Medford, MA}
  \country{United States}}
\email{tung.thai@tufts.edu}

\author{Ming Shen, Mayank Garg}
\affiliation{
  \institution{Arizona State University}
  \city{Tempe, AZ}
  \country{United States}}
\email{{mshen16, mgarg20}@asu.edu}

\author{Ayush Kalani, Nakul Vaidya}
\affiliation{
  \institution{Arizona State University}
  \city{Tempe, AZ}
  \country{United States}}
\email{{akalani2, nvaidya7}@asu.edu}

\author{Utkarsh Soni, Mudit Verma}
\affiliation{
  \institution{Arizona State University}
  \city{Tempe, AZ}
  \country{United States}}
\email{{usoni1, mverma13}@asu.edu}

\author{Sriram	Gopalakrishnan}
\affiliation{
  \institution{Arizona State University}
  \city{Tempe, AZ}
  \country{United States}}
\email{sgopal28@asu.edu}

\author{Neeraj Varshney, Chitta Baral}
\affiliation{
  \institution{Arizona State University}
  \city{Tempe, AZ}
  \country{United States}}
\email{{nvarshn2, chitta}@asu.edu}

\author{Subbarao Kambhampati}
\affiliation{
  \institution{Arizona State University}
  \city{Tempe, AZ}
  \country{United States}}
\email{rao@asu.edu}

\author{Jivko Sinapov}
\affiliation{
  \institution{Tufts University}
  \city{Medford, MA}
  \country{United States}}
\email{jivko.sinapov@tufts.edu}

\author{Matthias Scheutz}
\affiliation{
  \institution{Tufts University}
  \city{Medford, MA}
  \country{United States}}
\email{matthias.scheutz@tufts.edu}

\begin{abstract}

Learning to detect, characterize and accommodate novelties is a challenge that agents operating in open-world domains need to address to be able to guarantee satisfactory task performance. Certain novelties (e.g., changes in environment dynamics) can interfere with the performance or prevent agents from accomplishing task goals altogether. In this paper, we introduce general methods and architectural mechanisms for detecting and characterizing different types of novelties, and for building an appropriate adaptive model to accommodate them utilizing logical representations and reasoning methods. We demonstrate the effectiveness of the proposed methods in evaluations performed by a third party in the adversarial multi-agent board game Monopoly. The results show high novelty detection and accommodation rates across a variety of novelty types, including changes to the rules of the game, as well as changes to the agent's action capabilities.

\end{abstract}

\keywords{Open-world AI, Agent Architecture, Adaptive Multiagent Systems}

\newcommand{\BibTeX}{\rm B\kern-.05em{\sc i\kern-.025em b}\kern-.08em\TeX}

\begin{document}


\pagestyle{fancy}
\fancyhead{}


\maketitle 


\section{Introduction: Open-world AI}

Many classical adversarial AI tasks, such as game playing, take place in ``closed-world'' domains where all aspects of the domain---the types of entities, their properties, their actions, and the overall domain dynamics---are fixed. They are typically known to the agents before they start their task performance, and they do not change during task execution. Examples of such domains are ``perfect information games'' such as Chess, Go, or Ms.Pac-man, where the rules of the game, the goals of the players, and the entire state of the game are always known by all agents \cite{brown2018depthlimited,nash1,perez2016general}. This characteristic simplifies the game AI behavior by limiting the number of novelties to instances of known types (e.g., a chess move with the bishop a player has not seen before), thus allowing the development of the game AI without needing to anticipate any unknown scenarios within the bounds of the system (e.g., a novel piece with novel rules being introduced).

In contrast, agents operating in an ``open-world'' must be able to handle changes to entities and domain rules.  Specifically, in the context of open-world games, the rules, the state, and the actions of other players might only be partially known or could change anytime. The agent thus must discover these changes while playing the game \cite{boney2020learning,inbook}. Especially {\em interactive novelties} where agents interact with each other and with the environment present a challenge to any agent departing from a \textit{closed-world} assumption \cite{PONSEN200759,boardgame1,boardgame3}. In open-world environments, the action's effects and interaction's effects can be changed during the task operation time. Therefore, making the wrong move or wrongfully interacting with other agents can cause the agent to fail the task.     


In an effort to tackle the challenges of interactive novelties in adversarial open worlds, we propose general methods and architectural mechanisms that allow AI agents to detect, characterize, and adapt to interactive novelties in adversarial games. We develop a general novelty-handling framework, as well as symbolic logical reasoning methods to detect, learn, and adapt to novelties in \textit{open-world} environments. Our main contributions include (1) an architectural framework to handle interactive novelties in an adversarial environment, and (2) new logical reasoning approaches to characterize novelties and accommodate them during planning (that expands current state space, action space, and expected action effects).

\section{Background and Related Work}
Recent applications of multi-agent environments such as multiplayer games \cite{Peng}, Poker \cite{billings1998opponent}, social robotic systems \cite{Barakova}, and adversarial attack and defense \cite{REN2020346} consist of adversary elements and complex agent's behaviors. Therefore, learning how to adapt to the opponents' strategies becomes an essential task for current AI architecture. Unlike collaborative AI, where all the agents manage to work together to pursue a team goal, adversarial AI agents must learn other agents' behaviors to develop suitable strategies to maximize their own goals. This paper uses the open-world Monopoly environment as the primary test bed. Monopoly contains several main characteristics of an adversarial environment, such as unknown opponents' behaviors, stochastic elements (e.g., dice rolls, community cards, and chance cards), and novelties in the game. These characteristics can be found in many real-world domains, such as stock market forecasting, self-driving vehicles, or cybersecurity.

Current cognitive architecture systems such as probabilistic graphical models \cite{koller2009probabilistic,sucar2015probabilistic} provide an excellent tool that combines graph theory and probability theory to enable efficient probabilistic reasoning and learning. The model is widely used in the AI community as one of the main tools to generate state-of-the-art results. These results show the capabilities of the model to handle some of the challenges in traditional cognitive architecture, such as perception, interaction, and adaptation. However, these approaches are not explicitly developed to deal with \textit{closed-world} environment. Even though these methods have shown excellent results in \textit{closed-world} environments, addressing open-world and interactive novelty remains a challenge.

Over the past two decades, many research studies attempted to tackle the challenge of open-world AI. However, the challenge of integrating a general intelligence system capable of detecting and adapting to an open-world environment still remains unsolved \cite{DBLP:journals/corr/abs-2106-02204,AIBenGoertzel,Gizzi2021TowardCP,sarathy2020spotter}. Several challenges of integrating general AI systems are pointed out in previous studies, such as the difficulty of integrating the requisite capabilities (e.g., detecting novelty and adapting to the novelty), and the difficulty of measuring the performance of the agent towards human-like behavior AI \cite{AIBenGoertzel}. Reinforcement learning (RL) methods have been proposed as a solution for open-world environment \cite{Padakandla_2020,choi,ALA12-hester, arulkumaran2017deep} in recent years. These methods use past and present experience to learn a new representation of the world or attempt to construct a suitable control policy in dynamically-changing environments. However, RL and deep RL suffer to adapt to small environmental changes. Small pixels change in Atari arcade games can cause the RL agent to corrupt and fail to complete the task, and adaptation to novelties may often take as long as training the agent from scratch \cite{goelnovelgridworlds}. Finally, recent works in the explainable AI (XAI) literature have looked at answering contrastive queries \cite{sreedharan2020bridging}, which could very well be about potential novelties in the world. However, applying such a line of work for detecting open-world novelties would require an agent (assumed to be a human in the loop in XAI) to formulate and initiate queries to the agent to elicit the presence of novelties. Similarly, XAI works \cite{verma2022advice} that initiate an explanatory dialogue depending on the human in the loop (instead of automated detection and characterization) to analyze and detect open-world novelties. Finally, there are works that learn terms in the user's vocabulary \cite{soni2022preference}. The user can then use these terms to advise the agent on accommodating the novelty. 

Current approaches in cognitive AI systems such as Cognitive-Affective State System (CASS), and Sigma Cognitive Architecture have attempted to address the open-world AI challenge \cite{DBLP:journals/corr/abs-2101-02231,JiSigma,AIBenGoertzel}. Both architectures have been constructed to solve the problem without updating their core components or characterizing the novelty. These approaches may improve the overall performance of the AI. However, both architectures are not good enough to apprehend specific changes in the environment and accommodate those changes. More developments are needed for these architectures to perform well in a realistic open-world environment, where a part of the information can be changed, such as adversary mental models, transition functions, and agents' actions and interactions.   

\section{Preliminaries}
\subsection{n-Person Non-Cooperative Stochastic Turn-Based Games}
We consider $\mathcal{M} = \langle n, S, \{A_i\}_{i \in n}, T, R,
\gamma \rangle$ as the non-cooperative stochastic game environment
consisting of a finite, non empty state space $\mathcal{S}$; $n$
players, $\{1,2,\cdots,n\}$; a finite set of action set $\{A_1, A_2,
A_3, \cdots A_n \}$ for each of the players; a set of conditional
transition probabilities between states $T$, such that $T(s,
a_1,a_2,\cdots,a_n,s') = P(s'|s,a_1,\cdots,a_n)$; a reward function
$R$ so that $R:\mathcal{S} \times A \rightarrow \mathbb{R}$, where $A
= A_1 \times A_2 \times A_3 \times \cdots \times A_n$. An n-person
stochastic game is turn-based if at each state, there is exactly one
player who determines the next state. In order to formulate the
problem, we extend the action sets $A_i$ for $i \in \{1,2,\cdots,n\}$
to be state dependent. For each particular state $s$, there is a
restricted action set $A_{ir}$, there is at most $i \in
\{1,2,\cdots,n\}$ such that $|A_{ir}| > 1$.

At the beginning of the game, all players start at the same initial state $s_0 \in \mathcal{S}$. Each player independently performs an action $a_i^1 \in A$ . Given $s_0$ and the selected actions $a_1 = \{a_1^1, a_1^2,\cdots, a_1^n\} \in A$,  the next state $s_1$ is derived based on $s_0$ and $a_1$, with a probability $P(s_1|s_0,a_1)$. Then, each player independently performs an action $a_i^2$, next state $s_2$ is derived based on $s_1$ and $a_2$, with a probability $P(s_2|s_1,a_2)$. The game continues in this fashion for an infinite number of steps, or until the goal is reached. Therefore, the game generates a random history $h = \{s_0, a_1, s_1, a2,...\} \in H = S \times A \times S \times A ...$. Based on a partial of the history $h'= \{s_0, a_1, s_1, a2,...s_k\}$, we can derive the conditional distribution, so-called strategy $\pi_i(h') \in P(A_i)$, with $P(A_i)$ is the set of probability measures on $A_i$. A strategy set $ \pi = \{\pi_1; \pi_2,...,\pi_n\}$ consists of a strategy $\pi_i$ for each player $i$ is used to determine the next action $a_i^{k+1}$. Finally, the reward function $R$ is specified based on the transition of the game, and $\gamma \in (0,1]$ is the discount factor which determines the importance of immediate and long-term future rewards. 

\subsection{Interactive Novelty}
In general, novelty refers to a change in the environment where the agent can neither apprehend the change from its own knowledge base nor from its past experience. In this paper, we want to address the challenge of detecting and accommodating interactive novelty. More specifically, interactive novelty refers to the change in agent's actions, interactions, and relations. 
\begin{itemize}
    \item \textbf{Novelty Level 1 [Action}]: New classes or attributes of external agent behavior. 
    \item \textbf{Novelty Level 2 [Interaction}]: New classes or attributes of dynamic, local properties of behaviors impacting multiple entities.
    \item \textbf{Novelty Level 3 [Relation}]: New classes or attributes of static, local properties of the relationship between multiple entities.
\end{itemize}
We denote $\mathcal{C} = \{C_1, C_2, \cdots, C_n\} \in A$ as the interaction set of the agent. The set represents the agent's capability to interact with other agents, or with the environment. The relation set $\mathcal{L} = \{L_1, L_2, \cdots, L_n\} \in H$ represents the relationship of the agent with other agents, or agent with the environment, such that the relationship is shown as a part of the history, or action sequence. Each action $a_i$ in the action set $A$ is defined by a preconditions set $\delta_i(a)$ and an effects set $\beta_i(a)$. A preconditions set $\delta_i(a)$ of an action $a_i$ includes all the conditions that need to be satisfied in order to execute the action. Meanwhile, the effects set $\beta_i$ of an action $a_i$ indicates the expected results after a successful execution of action $a_i$.

The set of interactive novelties $\mathcal{N}$ consist of all the changes that can occur in action set, interaction set, and relation set. In this scenario, action novelty refers to changes in action space, action preconditions or action effects. We denote $A' = \{A_1', A_2', \cdots, A_n'\}$ as the new action set, which contains all unknown actions to the agent, such that $A' \cap A = \emptyset$, and $A' \notin \mathcal{KB}$, where $\mathcal{KB}$ is the agent knowledge base. We assume that the preconditions $\delta'$ and effects $\beta'$ of the new action set $A'$ are completely unknown to the agent, and both must be discovered through agent's interactions. Similarly, we can present the new interaction set as $\mathcal{C}'$, and relation set $\mathcal{L}'$, then formulate interaction novelty and relation novelty accordingly. 

\subsection{Problem Formulation: Interactive Novelty Detection and Adaptation}
The integrated architecture allows us to map all the essential information in \textit{section 3.1} of the environment to the knowledge base, $\mathcal{KB}$. Based on the information, we can construct the strategy $\pi$ using the truncated-rollout MTCS solver. However, because interactive novelties may occur throughout the course of the game, the plan must be adjusted in order to accommodate new actions, interactions, or relations.

As described in \textit{Section 3.1}, the pre-novelty environment is represented as a non-cooperative stochastic turn-based game: 
$$\mathcal{M} = \langle n, S, \{A_i\}_{i \in n}, T, R, \gamma \rangle$$
In order to detect and accommodate interactive novelties, we define a detection function $d(s,a,s')$ to determine if there is any unexpected change in the environment after the agent selects an action $a$ in state $s$ and observes the next state $s'$, or if the agent performed a new action. In addition, an identification function $\iota(s,a,s')$ characterizes the cause of the change based on logical reasoning. The purpose of these functions is to represent the new environment after novelty (post-novelty) $\mathcal{M}'$, such that $$\mathcal{M}' = \langle n, S', \{A_i'\}_{i \in n}, T', R', \gamma \rangle$$

\noindent where $S'$ is the new state space post-novelty. The set post-novelty $\{A_i'\}_{i \in n}$ is the new finite action set with respect to each agent $\alpha$ in the environment, $T'$ is the new conditional transition function, and $R'$ is the new reward function post-novelty. From the new model of the post-novelty world $\mathcal{M}'$, we modify the current strategy set $\pi$ in order to adapt to the changes in the environment, as described in the next section. 
 \section{Adversarial Domain: Open-World Monopoly} 
\subsection{Environment Implementation}
\begin{figure}[t]
        \centering
            \includegraphics[width=0.4\textwidth]{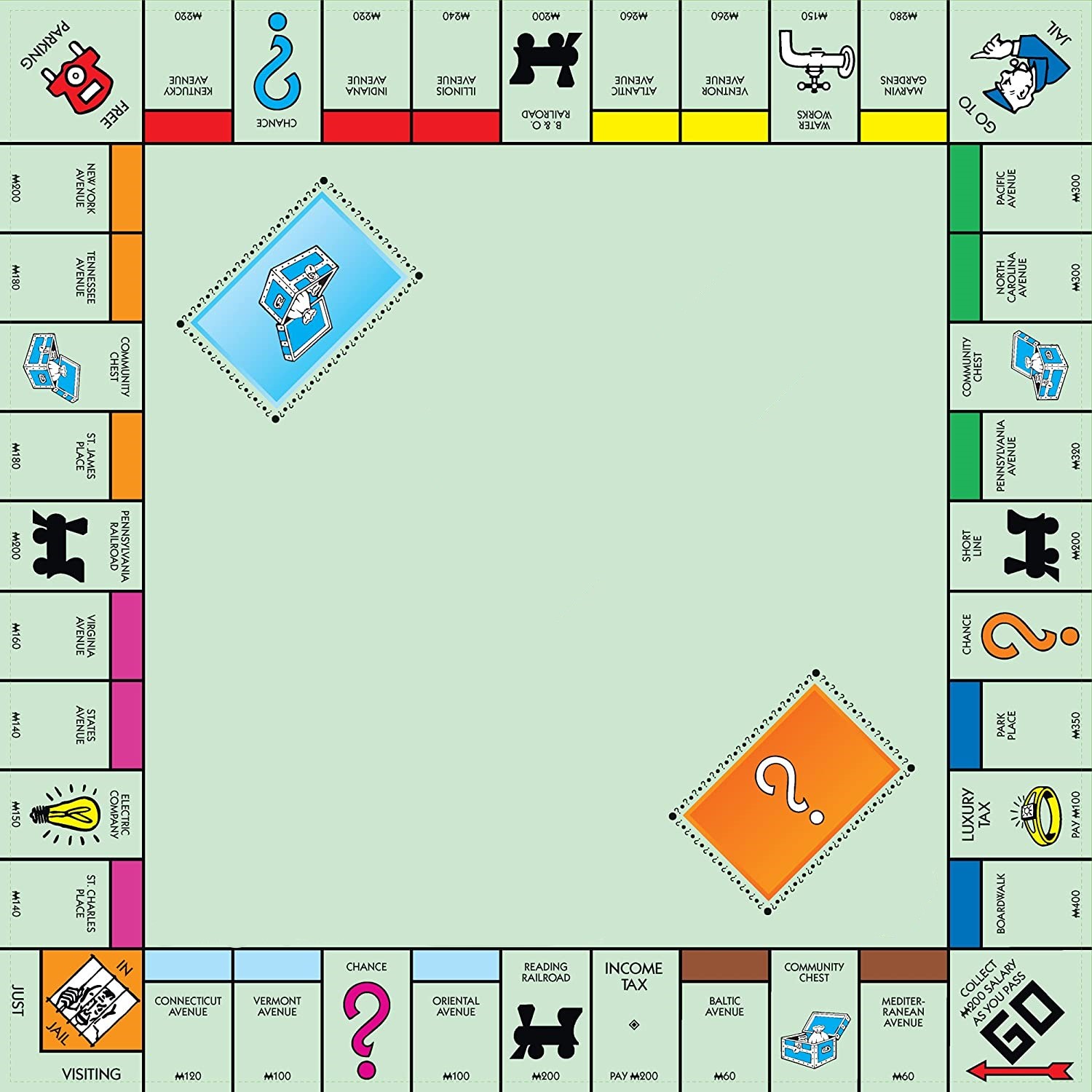}    
        \caption{Classic Monopoly Board}
        \label{board}
        \vspace{-1em}
    \end{figure}

Monopoly is a multi-player adversarial board game where all players start at the same position. The game can support up to 4 players, described in Figure \ref{board}. All players roll dice to move across the board. The ultimate goal of the game is to be the last player standing after bankrupting other players. This objective can be achieved by buying properties, and railroads, monopolizing color sets, and developing houses on properties. If one player lands on a property owned by another player, they get charged rent or a fee. After monopolizing color sets and developing houses and hotels, players can charge higher fees when other players land on their properties. The game includes different surprise factors such as chance cards, community cards, jail, auction, and trading ability between agents. These elements can completely change the game. Hence, any action in the game needs to be adapted to dice rolls, community cards, chance cards, and the decisions of other players. These game characteristics make it more challenging for integrated planning and execution. In the game simulator, novelties can be injected on top of the standard game to study how the agent detects and accommodates these changes \cite{mono}. The third-party team that ran the evaluation developed the Open-World Monopoly environment. Unlike traditional Monopoly, where we can fully observe all the states and actions of other agents, the Open-world Monopoly does not allow us to monitor all the actions and interactions on our turn \cite{KEJRIWAL2021102364}. So, the environment is partially observable. 

\subsection{Interactive Novelties in Monopoly}
\label{iNovelty}
We implement three different categories of interactive novelty discussed in \textit{Novelty Characterization} section into a classic Monopoly game. Some theoretical examples of novelty are described as below:
\begin{itemize}
    \item \textbf{Action Novelty}: This class of novelty can be illustrated through a stay-in-jail action. For this novelty, the player could stay in jail as long as they want. However, the player must pay a certain fee to the bank each time they receive rent (when the player decides to stay in jail by their own choice voluntarily).
    \item \textbf{Interaction Novelty}: We illustrate the relation novelty through a loan interaction between two agents. For example, a player could send a loan request to another player and pay the loan back over a specific amount of time that both parties agree. 
    \item \textbf{Relation Novelty}: We illustrate the relation novelty through a relation property, where we enforce a relation of homogeneity between properties in a specific monopolized color group (one color group). The player must homogeneously improve a monopolized set of properties in a given move. For example, imagine the player has 3 orange properties (a monopolized set). In the default game, you could set up a house on the first property, and leave the second one unimproved. For this novelty, in the move, if you improve the first property, you must also improve the second and third so that the properties are 'homogeneously' improved at the end of the move. Failure to do this will lead to the improvement being revoked at the end of the move.
\end{itemize}

\section{The Architectural Framework}

\begin{figure*}[t]
\centering
\includegraphics[scale=0.55]{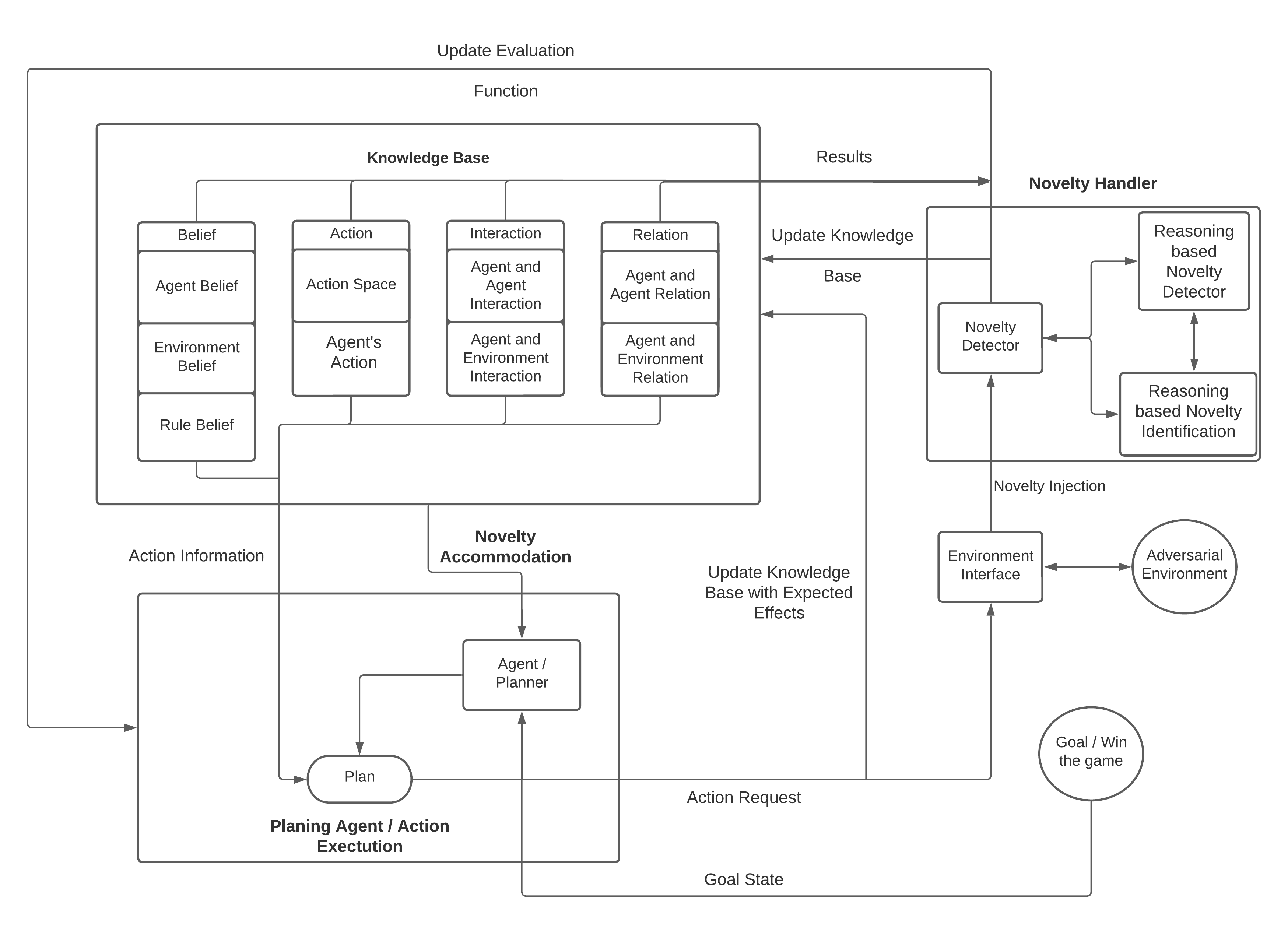}
\caption{The overall architecture of the novelty handling framework in an adversarial environment}
\label{architecture}
\end{figure*}

The architecture includes four main components: the environment
interface, the novelty handling component, a knowledge base, and a
planning agent, as shown in Figure~\ref{architecture}. The Novelty
Handler component was integrated in the ``Agent Development
Environment'' (ADE) \cite{Andronache2006AdeA} which allows for the
development of different integrated architectures. The Knowledge Base
of the agent is constructed by the \textbf{Belief Stack},
\textbf{Action Stack}, \textbf{Interaction Stack}, and
\textbf{Relation Stack}. The Planning Agent component develops and
operates the plan based on the information in the knowledge base and
the goal. The Monopoly Interface connects to the Monopoly API so that
novelties can be injected into the environment. These novelties are
detected and characterized by the novelty handler component. The
component is developed using Answer Set Programming (ASP), a
declarative programming oriented towards challenging search problems
\cite{baral_2003, baral-etal-2004-using,
  DBLP:journals/corr/GebserKKS14}. After the novelties are determined,
the novelty handler updates the new actions, effects, or states to the
knowledge base. When the agent receives the updated information, the
planning agent then reconstructs the plan according to the new
knowledge base.

\subsection{Novelty Detection}

We record the information of the game as provided by the game environment and compare it with our ``expectation'' state of the game board. This ``expectation'' state is derived from the agent's knowledge base of the game, including expected states, actions, action preconditions, and end effects. Then, the game environment provides us with the actual game board states and actions that have occurred between the current time step and the previous time our agent performed an action (e.g., after our agent lands on a property and buys it, all other agents get to act in order until it is our agent's turn again). When we notice a discrepancy between our expected state and the actual state, we surmise that something must have changed within the game, i.e., a novelty may have been introduced, which makes some aspects of our domain representation incorrect. Such unpredicted changes require the agent to update its knowledge base accordingly (e.g., a new action is added to the action space of Monopoly). An example of the novelty-handling component is shown in \textit{Algorithm 1}. The evaluation is run in a tournament setup (many games in a tournament), discussed in section \ref{results}. Therefore, when the agent detects a novelty from the current game, this novelty information will be used to adapt to the next game.
\begin{algorithm}[ht]
	\caption{Novelty Detection Pipeline} 
	\begin{algorithmic}[1]
	    \State Initialization: 
	        \State State space $\mathcal{S}$, Action space $\mathcal{A}$, Expected State $\mathcal{S}$'
	        \State $d(s_t,a_t, s'_t)$ = $False$
	        \State $\iota(s_t, a_t,  s'_t)$ = $None$
	        \State $t$ = 0 \Comment{Time step}
	   \While{Game does not end}
	        \If {$a_t = None$} \Comment{No action was performed}
	            \If {$S_{t+1} \neq S'_{t+1}$ }
	                \State $d(s_t, a_t,  s'_t)$ = $True$ \Comment{Novelty Detected}
	                \State $\iota(s_t, a_t,  s'_t)$ = $Relation$ \Comment{Novelty Characterization}
	            \Else 
	                \State $d(s_t,a_t, s'_t)$ = $False$
	                \State $\iota(s_t, a_t,  s'_t)$ = $None$
	            \EndIf
	        \Else     
	            \If {$a_t \notin \mathcal{A}$} \Comment{Unknown Action}
	                \State $d(s_t, a_t,  s'_t)$ = $True$ \Comment{Novelty Detected}
	                \State $\iota(s_t, a_t,  s'_t)$ = $Action$ \Comment{Novelty Characterization}\\

	            \Comment {\textit{Case 1: All precondition $\delta(a_{t+1})$ for action $a_{t+1}$ are satisfied but action $a_{t+1}$ is not executable}}
	            \ElsIf {$\delta(a_{t+1})$ == $True \land a_{t+1} == False$}
	               \State $d(s_t,a_t,  s'_t)$ = $True$ \Comment{Novelty Detected}
	               \State $\iota(s_t, a_t,  s'_t)$ = $Action$  \Comment{Novelty Characterization}\\
	               
	             \Comment {\textit{Case 2: At least one precondition for action $A_{t+1}$  is not satisfied but action $a_{t+1}$ is executable}}
	            \ElsIf {$\delta(a_{t+1}) == False \land a_{t+1} == True$}
	                    \State $d(s_t,a_t,  s'_t)$ = $True$ \Comment{Novelty Detected}
	                    \State $\iota(s_t, a_t,  s'_t)$ = $Action$  \Comment\Comment{Novelty Characterization}
	           \ElsIf
	                \State... \Comment{More Cases of Interactive Novelty}
	           \Else
	                \State $d(s_t,a_t,  s'_t)$ = $False$
	                \State $\iota(s_t, a_t,  s'_t)$ = $None$
	           \EndIf
	       \EndIf
	       \State t = t + 1
	\EndWhile \\
	\Return $d(s_t,a_t,  s'_t)$,  $\iota(s_t, a_t,  s'_t)$ 
	\end{algorithmic} 
\end{algorithm}

\subsection{Novelty Characterization}

Next, the agent uses a novelty identification module to characterize
the novelty. This module has several sub-modules (which can be run in
parallel), each focused on determining a specific novelty type. Each novelty identification sub-module uses the same ASP code
(except for two changes) that is used for hypothetical reasoning about the effect of an action. The first change is that, a particular parameter,
which is the focus of that specific sub-module, which was originally
a fact, is now replaced by ``choice'' rules of ASP that enumerate
different values that the parameter can take. The second change is
that constraints are added to remove possible answer sets where the
predicted game board state does not match the observed game board
state. The resulting program's answer sets give us the parameter values which reconcile the predicted game board state and the
observed game board state. If there is only one answer set and thus a
unique parameter value, then if this value is different from the value
we had earlier, we have identified a novelty. Now we can
update our ASP code that was used for hypothetical reasoning by simply
replacing the earlier value of the parameter with the new value.

Below we first give a glimpse of how ASP can be used for reasoning
about the next state and how that code can be minimally modified to
infer a novelty.

To reason about the next state, the ASP code will first define the game
parameters through facts such as the following:

\begin{verbatim}
dice_value(1..6).
player(player1;player2).
cash(1..1000).
asset("B&O_Railroad").
penalty(50).
\end{verbatim}

Then rules of the following form are used to define  actions and fluents. 

\begin{verbatim}
action(sell_property(P,X)) :- player(P), asset(X).
fluent(asset_owned(P,V)) :- player(P), asset(V).
\end{verbatim}

Properties of actions, such as their pre-conditions, and their effects are defined using rules of the following kind:

\begin{verbatim}
%Executability of selling assets
:- occurs(sell_property(P,V), T), player(P), 
   asset(V), time(T), not holds(asset_owned(P,V),T).

%Effect of selling assets
not_holds(asset_owned(P,V),T+1) :- 
              holds(asset_owned(P,V),T),
              occurs(sell_property(P,V),T),
              player(P), asset(V),time(T).
not_holds(asset_mortgaged(P,V),T+1) :- 
              holds(asset_owned(P,V),T),          
              occurs(sell_property(P,V), T), 
              player(P), asset(V), time(T).

holds(current_cash(P,X+Y),T+1) :- 
       holds(current_cash(P,X),T), 
       occurs(sell_property(P,V),T),
       not holds(asset_mortgaged(P,V),T),
       asset_price(V,Y), player(P), asset(V), time(T).
                                                        
not_holds(current_cash(P,X),T+1) :-  
       holds(current_cash(P,X),T), 
       occurs(sell_property(P,V),T),
        not holds(asset_mortgaged(P,V),T),
       asset_price(V,Y), player(P), asset(V), time(T).

holds(current_cash(P,X+Y),T+1) :-  
        holds(current_cash(P,X),T), 
        occurs(sell_property(P,V),T),
        holds(asset_mortgaged(P,V),T),
        asset_m_price(V,Y), player(P), asset(V), time(T).
                                                        
not_holds(current_cash(P,X),T+1) :-  
         holds(current_cash(P,X),T), 
         occurs(sell_property(P,V),T),
         holds(asset_mortgaged(P,V),T),
        asset_m_price(V,Y), player(P), asset(V), time(T).

%Executability of paying jail fine
:- occurs(pay_jail_fine(P), T), player(P), time(T), 
   not holds(in_jail(P), T).
:- occurs(pay_jail_fine(P), T), player(P), time(T), 
   not holds(current_cash(P, _), T).
:- occurs(pay_jail_fine(P), T), player(P), time(T), 
   holds(current_cash(P,X),T), X < 50.

%Effect of paying jail fine
not_holds(in_jail(P), T+1) :- holds(in_jail(P), T), 
              occurs(pay_jail_fine(P), T), 
              player(P), time(T).
not_holds(current_cash(P, X), T+1) :- 
  holds(current_cash(P,X),T), holds(in_jail(P), T), 
  occurs(pay_jail_fine(P), T), player(P), time(T).
holds(current_cash(P, X-50), T+1) :- 
  holds(current_cash(P,X),T), holds(in_jail(P), T), 
  occurs(pay_jail_fine(P), T), player(P), time(T).
\end{verbatim}

The inertia rules are expressed as follows:

\begin{verbatim}
holds(F,T+1) :-  fluent(F), holds(F,T), 
                 not not_holds(F,T+1), time(T).          
not_holds(F,T+1) :-  fluent(F), not_holds(F,T), 
                 not holds(F,T+1), time(T).   
\end{verbatim}

The initial state is defined using holds facts with respect to time step 0, such as:

\begin{verbatim}
holds(in_jail(player1), 0).
holds(current_cash(player1,500),0).
\end{verbatim}

An action occurrence at time step 0 is then defined as a fact in the following form.

\begin{verbatim}
occurs(pay_jail_fine(player1),0). 
\end{verbatim}

Now when a complete ASP program with rules and facts of the above kind is run, we get an answer set from which we can determine the state of the world at time step 1.

Suppose that the answer set has the facts:

\begin{verbatim}
holds(in_jail(player1), 0).
occurs(pay_jail_fine(player1),0).
holds(current_cash(player1,500),0).
holds(current_cash(player1,450),1).
\end{verbatim}

while our next observation gives us:

\begin{verbatim}
obs(current_cash(player1,477),1).
\end{verbatim}

The discrepancy between our prediction about player1's current\_cash being 500 (at time step 1) is different from our observation that player1's current\_cash is 477. This suggests there is a novelty. This can be determined by the following two simple rules. 

\begin{verbatim}
discrepancy(F,T) :- fluent(F), time(T), 
                    holds(F,T), not observed(F,T).   
discrepancy(F,T) :- fluent(F), time(T), 
                    not holds(F,T), observed(F,T).
\end{verbatim}

While the above could have been implemented in any language, including in the simulator language, which we also implemented in Python, having it in ASP, makes it easier for us to take the next step, which is to find out what the novelty is. 

In ASP, we have to modify the above ASP code by adding the following and removing ``penalty(50)'' (referring to the jail fine in the Monopoly game) from the original code.

\begin{verbatim}
oneto200(1..500).
1 { penalty(X) : oneto200(X)} 1.  %choice rule
:- obs(current_cash(P,X),1), 
   holds(current_cash(P,Y),1), X!=Y, player(P).
\end{verbatim}

In the above, the first fact and the choice rule defines the range of penalty that we are exploring. If we had just those two rules, we will multiple answer sets with a penalty ranging from 1 to 500. 
The constraint (the last ASP rule) then eliminates all the answer sets where the observation about current\_cash does not match with the holds. In the answer set that remains, we get the penalty value that would make the observation match with the holds, thus allowing us to figure out the novelty with respect to the penalty. In this particular case, the program will have the answer set with "penalty(23)" thus characterizing the novelty that the penalty is now 23.

\subsection{Novelty Accommodation}

Since novelties in the state (features, dynamics, actions) mean the
agent would have to replan often and would have to do so based on the
most updated information, we were interested in developing an online
planning algorithm to determine the best action. However, with
environments that are both \textit{long-horizon} and
\textit{stochastic}, using online planning approaches like Monte-Carlo tree search, quickly becomes intractable. To address
this problem, we formulate a truncated-rollout-based algorithm that
uses updated domain dynamics (learned from detected novelties) for a
few steps of the rollout and then uses a state evaluation function to
approximate the return for the rest of that rollout. In our evaluation
function, we use both domain-specific components and a more general
heuristic to approximate the return from the state after the truncated
rollout. Furthermore, to ensure the agent adapts to the detected novelties, we made both the environment simulator used for rollouts and the evaluation function sufficiently flexible and conditioned on the environment attributes; we only used a few tuned constants. Thus,
whenever a novelty was detected, we updated the relevant attributes in
our simulator and evaluation function before running our algorithm to
decide our actions. Using this approach, we are able to incorporate
novel information into our decision-making process and adapt
efficiently. An example of the whole process is shown in
\textit{Algorithm 2}.

We will now provide a detailed description of the rollout algorithm and the evaluation function. In our algorithm, when choosing the next best action in a given state, we execute multiple rollouts for each possible action and compute the mean return value for each action. Each rollout is terminated either when some terminal state is reached or when some $k$ number of actions have been taken. The rollouts use the updated domain dynamics of the environment. Due to potentially high branching factor, we keep these rollouts to be short (which also limits the effects of errors in our characterization of any novelties). 

However, to infer some approximation of the long-term value of an action, we use an evaluation function. Our evaluation function consists of two components: one that is domain-specific and the other that is heuristic-based and can be applied to any domain in general. The heuristic component of the evaluation function involves relaxing the domain and performing another rollout on the relaxed domain for some depth $l$. Some examples of relaxations include limiting adversarial agents' actions and determination of domain dynamics. For instance, in the case of the Monopoly domain, we prevent the agent from taking any buying or trading actions. On the other hand, the domain-specific component of the evaluation function computes the value of the state as the sum of two terms: $\mathcal{M}_\text{assets}$ and $\mathcal{M}_{\text{monopoly}}$ where $\mathcal{M}_\text{assets}$ is the value of all the assets the agent owns, whereas $\mathcal{M}_{\text{monopoly}}$ computes the maximum rent that the agent would get if it gains a Monopoly over any color, scaled down by how far the agent is to get the monopoly. 

\begin{algorithm}[ht]
	\caption{New Action Effect Novelty Handling} 
	\begin{algorithmic}[1]
	    \State Initialization: State Space S, Action Space A, precondition Set of all actions $\delta(o)$, 
	        \State $d(s_t,a_t)$ = $False$
	        \State $\iota(s_t, a_t)$ = $None$
	        \State $t$ = 0 \Comment{Time step}
	   \While{Game does not end}
	       \State Given $S_t \times a_t \overset{\beta(a_t)}{\rightarrow} S_{t+1}$ 
	            \If {[$\beta(a_t) \notin \beta(a)$] $\cup$ $(S_{t+1} \neq S_{t}')$}
	                \State $d(s_t,a_t)$ = $True$
	                \State $\iota(s_t, a_t)$ = $Action$
	            \Else 
	                \State $d(s_t,a_t)$ = $False$
	                \State $\iota(s_t, a_t)$ = $None$
	            \EndIf
	           \If {$\iota(s_t, a_t)$ = $Action$} 
	                \State{\textit{Update Action Set If Needed}}
	                \State $A.insert(\beta(a_t))$ 
	                \State{\textit{Update Action Effect Set If Needed}}
	                \State $\beta(a).insert(\beta(a_t))$ 
	                \State{\textit{Update Action Precondition Set If Needed}}
	                \State $\delta(a).insert(\delta(a_t))$ 
	                \State $\mathcal{M}_{assets} \leftarrow \mathcal{M}'_{assets}$ \Comment{Update assets value}
	                \State $\mathcal{R}_{s} \leftarrow \mathcal{R}'_{s}$ \Comment{Update short-term gain}
	                \State $\mathcal{R}_{l} \leftarrow \mathcal{R}'_{l}$ \Comment{Update long-term gain}
	                \State $\mathcal{M}_{Monopoly} \leftarrow \mathcal{M}'_{Monopoly}$ \Comment{Update Monopoly beneficial value}
	           \EndIf
		    \State t = t + 1
		\EndWhile \\
	\Return $ND$
	\end{algorithmic} 
\end{algorithm}

\section{Evaluation \& Results}
\label{results}
\subsection{External Evaluation}
In an effort to maintain the integrity of the evaluation, all the information about the novelty was hidden from our team, and all the information about our architecture or methodologies was also hidden from the evaluation team. The external evaluations were performed by a third-party team that originally created the Open-world Monopoly domain. Our agent was evaluated on the three interactive novelties: \textit{action}, \textit{interaction}, and \textit{relation}. For each type of interactive novelties, more than 20 different novelties were introduced during the evaluation process. Each novelty level also has three difficulty levels (shown in \textit{Table 2}). The difficulty levels expressed the difficulty of detecting and using the novelty. For instance, if the novelty involved an action, an \textit{easy} difficulty means the action was available for the agent to perform without any preconditions. A \textit{medium} difficulty means the actions can be detected by observing other agents. A \textit{hard} difficulty means the agent can only act under specific circumstances, and it may require the agent to explore the environment to learn the action. There were more than 60 novelties in which the agent was evaluated in total. At least 600 tournaments (100 games per tournament) were run to measure our agent's performance. Tournaments were started with a traditional Monopoly game (no novelty). At a certain point throughout the tournament (non-specific, e.g., on the $5^{th}$ game), a novelty was introduced. To avoid ambiguity between novelties, only one specific novelty at a time was injected into a tournament. In our internal evaluation, ASP performed excellently in novelty detection and characterization. However, due to the characteristics of ASP, the novelty-handling component run time can be very high. Moreover, due to the nature of the game Monopoly (the game can go on indefinitely), and limited computational resources, we decided to use Python to overload the requirements of our solver and leverage the access to the simulator instead of using ASP to the first model and subsequently detect for novelties, to optimize the run time for the external evaluation.

Our agent was evaluated based on four different metrics. M1 is the percent of correctly detected trials (CDT). In this case, the percent of CDT is the percent of trails that have at least one True Positive and no False Positives (FP). M2 is the percent FP (the agent reports novelty when no novelty exists). M3 is the novelty reaction performance (NRP) before the novelty was introduced (pre-novelty). M4 is the novelty reaction performance
(NRP) after the novelty was introduced (post-novelty). To measure the NRP, our agent was evaluated against a heuristic agent which embedded some of the most common strategies in Monopoly (e.g., target some specific color, never buy some properties, always reserve money, etc.). Finally, we compute the novelty reaction performance (NRP) of the agent based on the following formula:
$$ NRP = \frac{\mathcal{W}_{agent}}{\mathcal{W}_{baseline}} $$
Where, $\mathcal{W}_{agent}$ is the win rate of our agent (pre-novelty for M3, and post-novelty for M4). $\mathcal{W}_{baseline}$ is $65\%$. 
\begin{center}
    \begin{table}[t]
        \begin{tabular}{ |p{1cm}||p{2cm}| p{2cm}| p{2cm}| }
             \hline
               \multicolumn{4}{|c|}{Novelty Level 1: Action} \\
             \hline
             Metrics & Easy & Medium & Hard \\
             \hline
                & Mean  & Mean  & Mean \\
             
             \hline
             M1 & $100\% $ & $100\% $ & $100\% $ \\
             M2 & $0\% $ & $0\% $& $0\% $\\
             M3 & $ 141.54\% $ & $ 129.23\% $ & $ 136.92\% $\\ 
             M4 & $ 151.79\% $ & $ 135.38\% $ & $ 143.08\% $\\
             \hline
               \multicolumn{4}{|c|}{Novelty Level 2: Interaction} \\
             \hline
             M1 & $100 \% $ & $100\% $ & $100\% $ \\
             M2 & $0\% $ & $0\% $& $0\% $\\
             M3 & $ 124.31\% $ & $ 142.77\% $ & $ 121.85\% $\\ 
             M4 & $ 130.46\% $ & $ 134.15\% $ & $ 113.23\% $\\
             \hline
               \multicolumn{4}{|c|}{Novelty Level 3: Relation} \\
             \hline
             M1 & $100\% $ & $100\% $ & $80\% $ \\
             M2 & $0\% $ & $0\% $ & $0\% $\\
             M3 & $ 147.08\% $ & $ 132.31\% $ & $ 150.15\% $\\ 
             M4 & $ 146.46\% $ & $ 121.85\% $ & $ 145.23\% $\\
             \hline
        \end{tabular}
        \caption{Evaluation}
        \vspace{-3em}
    \end{table}
    \vspace{-1.5em}
\end{center}

The results suggest that our cognitive architecture provides outstanding solutions for the game regardless of the complexity of the environment and differing levels of novelty. Furthermore, our agent achieved a perfect precision score ($100\%$ in percent of CDT and $0\%$ of FP) at all difficulty levels of action and interaction novelties. The agent achieved a nearly perfect precision score in relation novelties. However, the agent missed $20\%$ of the novelties in the hard level of difficulty. These failures to detect certain novelties happened due to the nature of the relation novelty category: we can only detect these novelties types when a specific action is executed. Due to the stochasticity of the Monopoly game, the agent would sometimes not perform a specific action throughout the entire evaluation. To identify a relation novelty, the agent may need to perform a particular action at a specific state to reveal the novelty. For example, in the relation property novelty scenario (discussed in section \ref{iNovelty}), this novelty only occurs when we monopolize a green color group (one of the most challenging color groups to monopolize due to the cost of each property). The agent may then fail to detect the novelty because the agent would never monopolize the green color group throughout testing. The M3 and M4 NRP scores in all novelty levels show that our agent outperformed the baseline agent before and after when novelties were introduced. The scores in \textit{Table 1} indicate that our cognitive architecture and accommodation strategies allow the planning agent to handle interactive novelties perfectly.   

\subsection{Internal Evaluation}
\subsubsection{Agent Performance Without Novelty Accommodation}
In order to understand the effectiveness of the novelty handler components (detection, characterization, and accommodation), we conduct experiments on all the novelties and record the win rate of the MCTS agent with and without the support from the novelty handler across all the novelties with a random number of games for each game tournament. Table \ref{table1} shows the overall performance of the MCTS agent with the novelty handler against the MCTS agent without the novelty handler. The result suggests that the MCTS agent with the support of the novelty handler outperforms the vanilla MCTS agent without novelty handling. There is a significant $10\%$ win rate difference between them. Furthermore, the results also indicate that the novelty handler components play an essential role in the agent's performance in an adversarial open-world domain. Although some novelties can have an essential effect on the game, and some novelties may not affect the game (nuisance novelties), the novelty handler mechanism still shows its efficiency in enhancing the agent's performance. For example, restricted color novelty can significantly affect the agent's strategies for buying and trading properties. On the other hand, other novelties, such as selling houses or property rates, can have minimal effects on the game.
\begin{table}[H]
    \begin{tabular}{|p{2cm}||p{2.5cm}|p{2.5cm}|} 
         \hline
         Novelty & Win rate of adaptive MCTS agent & Win rate of non-adaptive MCTS agent\\
         \hline
            & Mean $\pm$ SD & Mean $\pm$ SD \\
         \hline
         Action & $83.22\% \pm 5.33\%$  & $76.38\% \pm 6.313\%$ \\ 
         \hline
         Relation & $81.86\% \pm 7.26\%$  & $68.45\% \pm 5.164\%$ \\ 
         \hline
         Interaction & $89.6\% \pm 9.01\%$  & $72.5\% \pm 7.692\%$ \\ 
         \hline
    \end{tabular}
    \caption{Evaluation Results Of Agent's Performance With and Without Novelty Handler}
    \label{table1}
\end{table}
\subsubsection{Agent Performance Against Existing Methods}
In order to learn the performance level of our agent, we compare our agent against other Monopoly-playing agents. For this experiment, we evaluate our agent's performance against the hybrid deep reinforcement learning agent \cite{Bonjour2021DecisionMI}. The hybrid agent uses proximal policy optimization (PPO) and double-deep Q-learning (DDQN) algorithms. The authors compare the standard reinforcement approach to their hybrid approach, and the experimental results show that the hybrid agents outperform traditional RL agents. Significantly, the hybrid PPO agent has a win rate of $91\%$ against a fixed-policy agent that is developed base on the Monopoly world champion's strategy. In our evaluation, we ran two instances of our agents against one of the fixed-policy agents and the hybrid deep reinforcement learning agent in trials. The results are shown in table \ref{table2}. The results show our agent's dominant performance against the hybrid reinforcement learning approach. Our agent has a more than $85\%$ win rate in the tournament compared to $12\%$ of the hybrid learning agent.
\begin{table}[H]
    \begin{tabular}{|p{1.3cm}||p{1.8cm}|p{1.8cm}|p{1.8cm}|} 
         \hline
          & Our Agent 1 & Our Agent 2 & Hybrid Agent\\
         \hline
            & Mean $\pm$ SD & Mean $\pm$ SD & Mean $\pm$ SD  \\
         \hline
         Win ratio & $43.65\% \pm 1.42\%$  & $42.43\% \pm 1.33\%$ & $12.13\% \pm 1.51\%$\\ 
         \hline
    \end{tabular}
    \caption{Evaluation Results Of Agent's Performance Against The Hybrid Deep Reinforcement Learning Agent}
    \label{table2}
\end{table}
\section{Conclusion}

Our work presented a new agent architecture for interactive novelty handling in an adversarial environment that can detect, characterize, and accommodate novelties. Our architecture is modeled based on the thought process of human cognition when we deal with environmental changes. First, we use ASP to detect and characterize interactive novelties (action, interaction, and relation). Then, we update the detected novelties to our agent's knowledge base. Finally, we utilize the truncated-rollout MCTS agent to accommodate the novelty. The external evaluation results support the cognitive architecture's effectiveness in handling different levels of interactive novelty. 

However, the architecture has potential limitations in novelty characterization and learning agent behavior. One limitation of this architecture is the capability to learn the opponents' behaviors. Our cognitive architecture does not explicitly model the opponent's strategy to detect the change in other agents' behaviors and adapt accordingly. To address this limitation, we propose two additional models that can be a part of the novelty handler component. The first approach is to model the opponents' behavior using probabilistic reasoning \cite{PEARL198877, 1988i, JP1988ix}. In these models, we can learn the action probability distribution based on the game state, which helps us detect any change in opponents' behaviors. 

Secondly, we would like to model the opponents' behavior using reinforcement learning. Recent applications of reinforcement learning show promising results in learning opponents' behavior without knowing opponent's observations and actions during both training and execution processes \cite{DBLP:journals/corr/abs-2001-10829,DBLP:journals/corr/abs-2011-07290}. Ultimately, we believe improving the model's capability of predicting another agent's behaviors is the biggest area for growth.


\begin{acks}
This work was funded in part by DARPA grant W911NF-20-2-0006. We would like to thank Mayank Kejriwal, Shilpa Thomas, Min-Hsueh Chiu and other members of the University of Southern California team for the Monopoly simulator and agent evaluation.
\end{acks}


\bibliographystyle{ACM-Reference-Format} 
\bibliography{sample}


\end{document}